\documentclass{article}

\usepackage{PRIMEarxiv}
\usepackage{amsmath}
\usepackage[utf8]{inputenc} 
\usepackage[T1]{fontenc}    
\usepackage{url}            
\usepackage{booktabs}       
\usepackage{amsfonts}       
\usepackage{nicefrac}       
\usepackage{microtype}      
\usepackage{lipsum}
\usepackage{fancyhdr}       
\usepackage{graphicx}       
\graphicspath{{media/}}     
\usepackage{float}
\usepackage[colorlinks=true, linkcolor=black, citecolor=black, urlcolor=blue]{hyperref}

\title{Finetuning YOLOv9 for Vehicle Detection: Deep Learning for Intelligent Transportation Systems in Dhaka, Bangladesh
}

\author{
  Shahriar Ahmad Fahim\\
  Department of Transdisciplinary Science and Engineering \\
  School of Environment and Society\\
  Tokyo Institute of Technology \\
  Tokyo, Japan\\
  \texttt{fahim.s.aa@m.titech.ac.jp} \\
}

\begin{document}
\maketitle

\begin{abstract}
Rapid urbanization in megacities around the world, like Dhaka, has caused numerous transportation challenges that need to be addressed. Emerging technologies of deep learning and artificial intelligence can help us solve these problems to move towards Intelligent Transportation Systems (ITS) in the city. The government of Bangladesh recognizes the integration of ITS to ensure smart mobility as a vital step towards the development plan "Smart Bangladesh Vision 2041", but faces challenges in understanding ITS, its effects, and directions to implement. A vehicle detection system can pave the way to understanding traffic congestion, finding mobility patterns, and ensuring traffic surveillance. So, this paper proposes a fine-tuned object detector, the YOLOv9 model to detect native vehicles trained on a Bangladesh-based dataset. Results show that the fine-tuned YOLOv9 model achieved a mean Average Precision (mAP) of 0.934 at the Intersection over Union (IoU) threshold of 0.5, achieving state-of-the-art performance over past studies on Bangladesh-based datasets, shown through a comparison. Later, by suggesting the model to be deployed on CCTVs (closed circuit television) on the roads, a conceptual technique is proposed to process the vehicle detection model output data in a graph structure creating a vehicle detection system in the city. Finally, applications of such vehicle detection system are discussed showing a framework on how it can solve further ITS research questions, to provide a rationale for policymakers to implement the proposed vehicle detection system in the city.

\end{abstract}


\section{Introduction}
Intelligent transportation systems (ITS) field has emerged due to the rising  mobility demand and its consequences, such as traffic congestion, inefficiencies, and accident risks in transportation \cite{dimitrakopoulos2010intelligent,data2020ieee}. Rapid urbanization with population growth has affected cities around the world with these problems \cite{jiang2024intelligent}, and Dhaka of Bangladesh is not an exception. The megacity has been facing extreme traffic congestion \cite{haider2021cost} due to lack of adequate transportation infrastructure, unregulated vehicles, and poor traffic management. Researchers have recommended to improve the public transportation infrastructure, traffic management, traffic flow, and safety analysis through cutting-edge intelligent systems \cite{ali2023traffic}. 

Artificial intelligence (AI) has gained a lot of limelight recently, as it has made revolutionary impacts in every sector, including intelligent transportation. One of the reasons is the recent successes of deep learning techniques, a subset of AI \cite{lecun2015deep}. Utilizing these techniques is key in realizing ITS in society \cite{harle2024advancements}. AI and deep learning based solutions can help in decision support of urban traffic management to solve traffic congestion, monitoring \cite{hong2020traffic,lin2022intelligent}, and analysis in ways that cannot be done through traditional investment in road hardware \cite{li2023construction}. By detecting vehicles on congested roads, near real-time inference and decision-making may be done using object detection algorithms. These deep learning based vehicle detection algorithms are built on usually YOLO (You Only Look Once), SSD, R-CNN, Faster R-CNN, etc. Among them, YOLO and SSD are one-staged detectors and the rest are two-staged. Although two-staged detectors sometimes exhibit higher accuracy but come at a cost of increased computational expense \cite{ma2024review}. YOLO has a higher advantage over its competing algorithms in vehicle detection, Faster R-CNN and SSD because of better speed of inference and better detection of smaller objects, respectively \cite{zhao9review,zhao2022improved,kim2019multi}.

The YOLO model was first introduced by \cite{redmon2016you} in 2016. The paper approached the object detection as a regression problem, starting with pixels of images and then predicting bounding box and class probabilities. This approach improved the object detection accuracy and speed. Since then, a family of newer versions have been proposed \cite{redmon2017yolo9000,redmon2018yolov3,bochkovskiy2020yolov4,jocher2020yolov5,li2022yolov6,wang2023yolov7}, improving on the top of the predecessor models, and the latest addition being YOLOv9 \cite{wang2024yolov9}. It is the current state-of-the-art YOLO model with improvements of 0.6\% AP from the previous version (YOLOv8 \cite{Jocher_Ultralytics_YOLO_2023}). Although the general YOLO models are trained and bench-marked on general purpose datasets, such as MS COCO dataset \cite{lin2014microsoft}, these models have been consistently modified, fine-tuned, and applied to specific detection tasks in various domains by many researchers \cite{hussain2024yolov1,chien2024yolov9,talaat2023improved,zhou2023rde,shokri2023comparative,fang2019tinier,li2021yolo,baoyuan2021research}, including vehicle detection \cite{zarei2022fast,mahto2020refining,zhang2022real,omari2023optimizing,al2023vehicle,li2023modified,wang2020data,carrasco2021t,zhao2022improved}. These models can be deployed in cameras for real-time vehicle detection.  After the latest YOLOv9 model has been proposed \cite{wang2024yolov9,chang2023yolor}, and there has been no work implementing the state-of-the-art YOLO model for the vehicle detection task to our knowledge. Furthermore, no work has yet finetuned the latest YOLO version to a Dhaka, Bangladesh based dataset. Therefore, we pursue this objective.


The need to finetune the model for a region-specific dataset is, Dhaka or other cities in Bangladesh, have unique classes of vehicles that have their own features. They are often dissimilar from vehicles of other countries. So, we need to develop a model that is learned on the features of local vehicle types. In the context of Bangladesh, there have been some studies that aimed to do so. In 2022, after comparing different variations of YOLO models, researchers in \cite{alamgir2022performance} argued that YOLOv6 and YOLOv7 may emerge as better performing models for vehicle detection compared to the previous ones. A contribution by \cite{sarker2023bangladeshi} created a dataset with 20 classes (not publicly available) and found that YOLOv7 performed better than v5 and Scaled v4 with mAP of 0.802. Another recent contribution by \cite{munir2023vehicle} implemented a YOLOv7 based vehicle detection in Dhaka using Dhaka-AI dataset \cite{DVN/POREXF_2020}. Researchers in \cite{hossain2023real} used the YOLOv7 to detect eight types of vehicles commonly moving on the Padma Multipurpose Bridge from their own custom dataset (not publicly available) in Bangladesh and showed 0.981 mAP for using in toll collection to realize ITS. Another work \cite{salekin2023bangladeshi} refined YOLOv8 for vehicle detection in Dhaka using Poribohon-BD dataset \cite{tabassum2020poribohon} and obtained an mAP of 0.913. Currently in literature, no work can be found that used an object detection model other than YOLO variants such as SSD, RCNN, etc. for vehicle detection on a Bangladesh-based dataset. Results of recent works on vehicle detection will be compared later with our model in the Discussions \ref{subsec:comparison}. It is generally observed that the latest model exceeds the performance of the previous one, both by speed and accuracy, one of the motivations for performing the current research. Regarding implementing such vehicle detection practically in Bangladesh, although there is some recent effort in developing a real-time vehicle detection system in Dhaka \cite{dailystar2024}, a widespread implementation is yet not achieved. One underlying reason is the lack of understanding of the benefits of such vehicle detection system across the city. Smart mobility and ITS in transportation is one of the milestones of realizing the "Smart Bangladesh Vision 2041", an agenda for development by the Government of Bangladesh \cite{pal2023smart}. In 2019, a report published by Dhaka: Government of the People's Republic of Bangladesh identified understanding ITS and its impacts, policy, and strategy, as key challenges among others for ITS implementation in the country \cite{karim2019its}. None of the previous studies based in Bangladesh have discussed how a vehicle detection model can be used as a system to solve various transportation challenges. So, we also explore the uses of our model in society and how it can take us toward intelligent transportation systems.

\begin{figure}[t]
\centering
\includegraphics[width=3in]{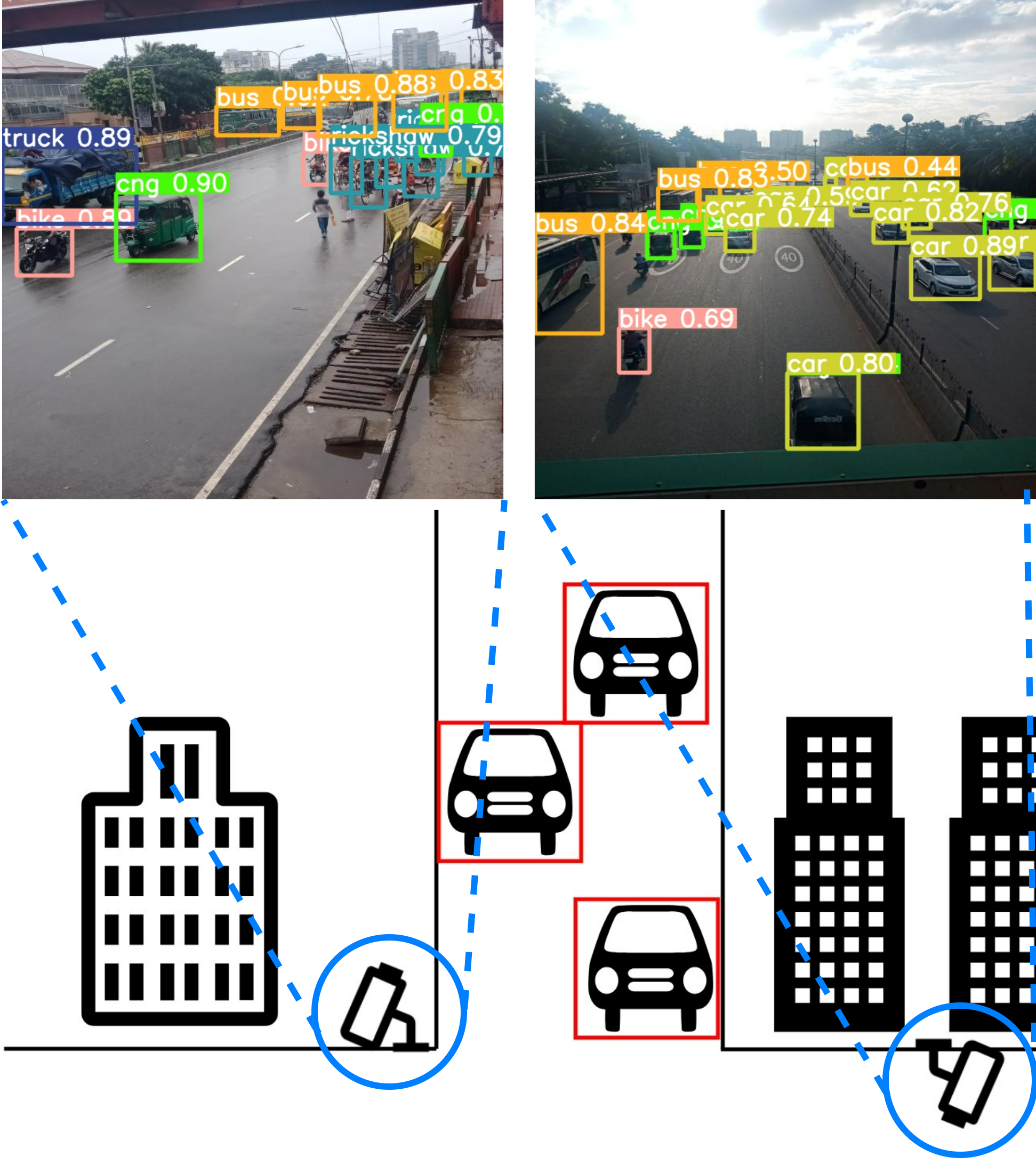}
\caption{Our proposed finetuned YOLOv9 model set up in cameras in the city to detect vehicles real-time. Analyzing the results of such vehicle detection system will assist in decision and policy-making to solve transportation challenges.}
\label{fig_abstract}
\end{figure}

In this paper, we implement the YOLOv9 model by fine-tuning on a dataset based in Bangladesh. The model should be deployed to establish a vehicle detection system in the city by deploying the model through CCTV. With the models deployed in cameras throughout the city, we propose a conceptual example technique to process the detection model output data to be represented in a graph structure data with respect to roads in the city. We discuss future directions on the top of the proposed vehicle detection system in Dhaka, Bangladesh that can solve various research questions to tackle transportation challenges in ITS through better policy and decision-making. Such discourse not being available in the literature, will help policy-makers to understand the impacts of a widespread vehicle detection system in the city (see Figure \ref{fig_abstract}). 

\newpage

Key contributions of our paper are as follows.

\begin{itemize}
    \item Fine-tuned the latest YOLOv9 model on a dataset based in Bangladesh for detecting vehicles, achieving a state-of-the-art performance of 0.934 mAP at IoU of threshold 0.5
    \item Proposed a conceptual technique to represent the model output data into graph structure for further algorithm developments if deployed through CCTVs or cameras in the city
    \item Discourse on potential applications of the proposed vehicle detection model to give a more satisfactory understanding of its benefits for intelligent transportation systems in the city, as as rationale for implementation and future directions
\end{itemize}

\section{Methods and Experiments}
\label{sec:methods}
\begin{figure*}
\centering
\includegraphics[width=6.2in]{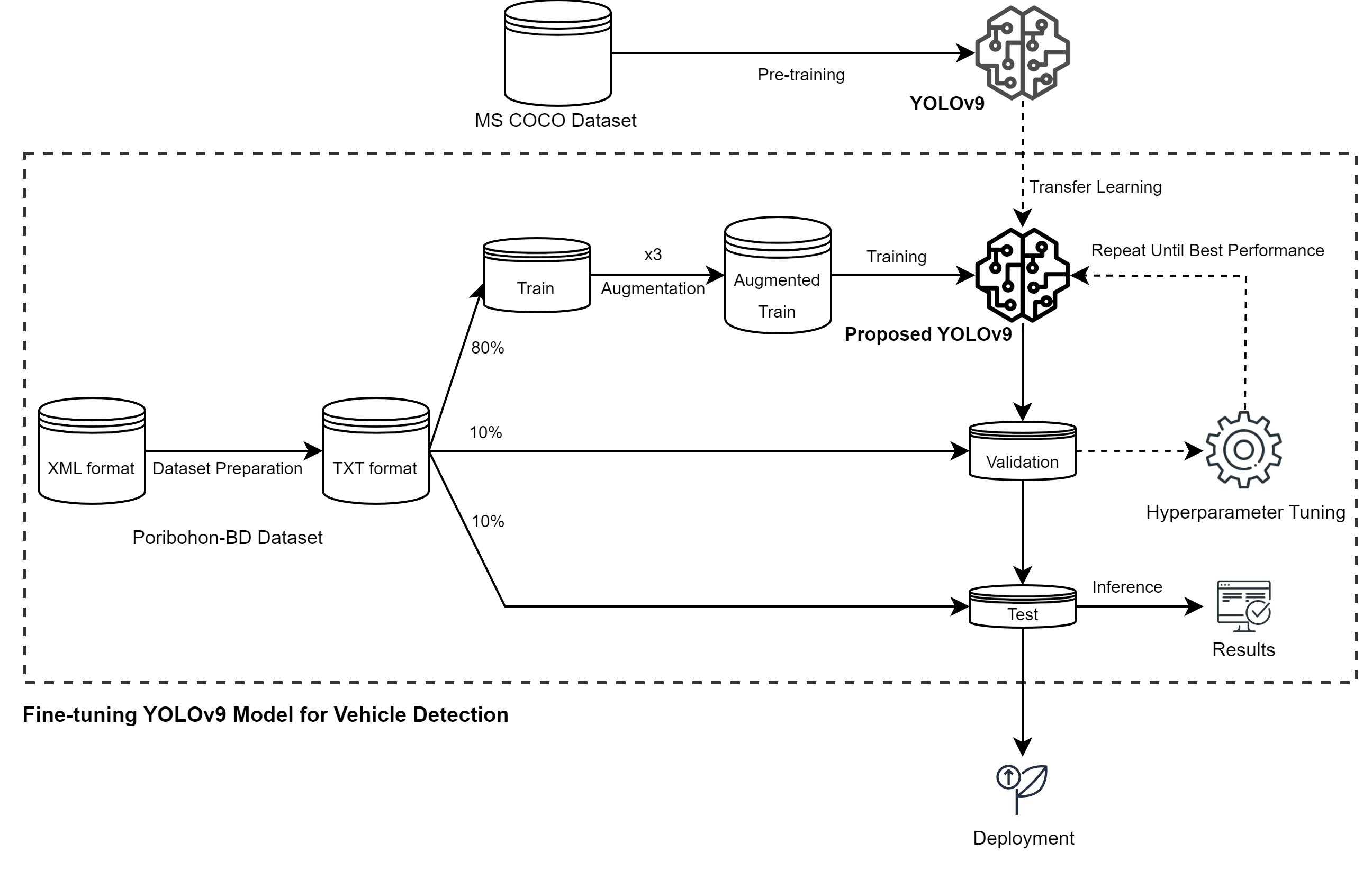}
\caption{Schematic of our methodology for fine-tuning the YOLOv9 model for vehicle detection. The pre-training is the training of the model in the original work of YOLOv9 \cite{wang2024yolov9}. The weights are imported as pre-training weights. Only the methodology inside the dashed line is implemented, so deployment is out of scope of this paper.}
\label{fig_methodology}
\end{figure*}

We present a schematic of our methodology through Figure \ref{fig_methodology}.

\subsection{Dataset Preparation}\label{subsec_dataset}

\begin{figure}
\centering
\includegraphics[width=3.5in]{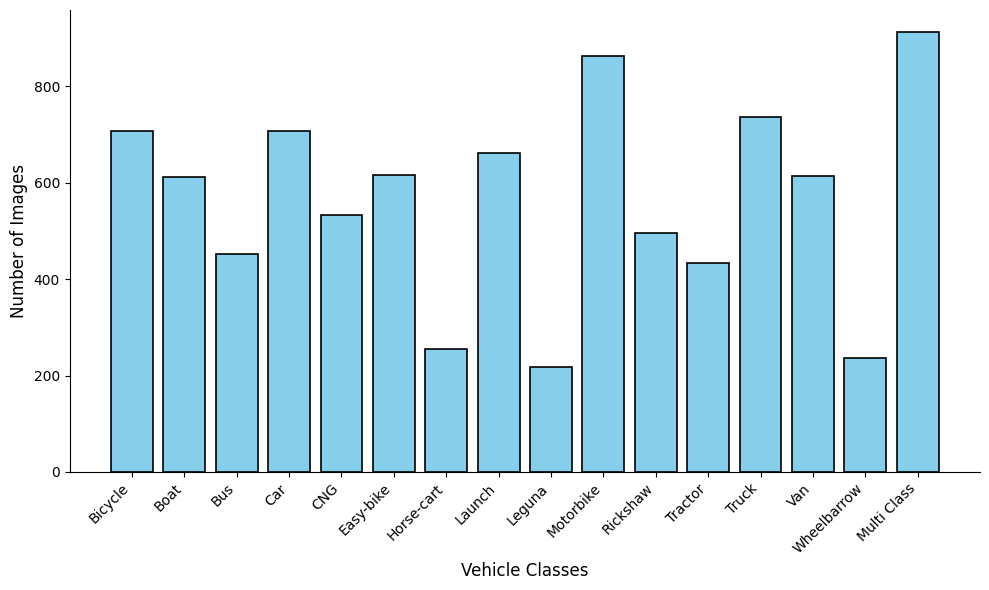}
\caption{Class balance of Poribohon-BD dataset \cite{tabassum2020poribohon} showing number of images for each type of vehicle.}
\label{fig_poribohonbd}
\end{figure}

The dataset we have chosen for this experiment is the Poribohon-BD dataset \cite{tabassum2020poribohon} because it is large in size compared to other publicly available datasets based in Bangladesh. It has a total of 9058 labeled and annotated images of 15 types of native Bangladeshi vehicles. Human faces were blurred for privacy reasons. The dataset consists of images with variance in pose, view, and angle of the cars. Different lighting conditions such as, day (sunny), night, and low light environments as well as some weather conditions such as rainy are present in the data. They were collected from from the highways of Bangladesh, where 1791 images were produced by augmentation which is almost one-fifth of the whole dataset, and 4000 images were collected from Facebook (social media). The dataset has a total of 16 folders (classes) where 15 are unique classes and the last is mixed-class vehicles. The folders of each vehicle class do not necessarily mean annotations of only that particular class exist inside that folder. Figure \ref{fig_poribohonbd} shows the number of images of each vehicle class in the Poribohon-BD dataset. The number of images as well as the number of times each vehicle class appear throughout the dataset varies largely as some vehicle classes are more frequently seen on the road than others. 

Originally, the dataset images were labeled with class names, and annotated values were provided in XML format (PASCAL VOC format). It was converted to PyTorch TXT format (YOLOv9 format) using an end-to-end computer vision platform called Roboflow \cite{Dwyer2024}, currently widely used by many computer vision researchers and engineers. A version of the dataset was generated by applying auto-orient and resizing to 640x640. Train-validation-test split was 80-10-10\%. Three images per training example was produced through augmentations: rotation, shear, saturation, brightness, exposure, and bounding box shearing.


\subsection{YOLOV9 Network Architecture}

Neural networks are susceptible to information loss in deep networks, a phenomenon known as information bottleneck \cite{tishby2015deep}. This leads to biased gradient flows, resulting degradation in accuracy. Researchers of YOLOv9 \cite{wang2024yolov9} proposed a new concept called programmable gradient information (PGI) to solve this issue. There are three components in PGI - (1) main branch, (2) auxiliary reversible branch, and (3) multi-level auxiliary information. The auxiliary reversible branch solves the information bottleneck problem in the main branch of deep networks and is applicable to shallower networks as well. It is not required during inference time, so does not increase inference cost. The multi-level auxiliary information reduces the error that is accumulated during training deep networks. YOLOv9 is proposed by combining  PGI with another concept called Generalized Efficient Layer Aggregation Network (GELAN). GELAN is again a combination CSPNet \cite{wang2022designing} and, ELAN \cite{wang2020cspnet}. GELAN is suitable as a lightweight architecture with competitive inference speed and accuracy. Only the GELAN model (without PGI) is also released with YOLOv9 as an alternative, a reasonable choice for a training environment with a low computational power. It should be mentioned that both YOLOv9 and GELAN have similar corresponding model parameter sizes during inference, but GELAN is lighter during training without PGI, resulting decreased performance compared to YOLOv9. By solving the gradient flow issue and information loss, we argue that YOLOv9 architecture can effectively improve the vehicle detection performance of neural network models. 

Four different variants for YOLOv9 (S: small, M: medium, C: compact, and E: extended) in order of architecture size are introduced in the original paper, along with four variants GELAN, while their GitHub repository also mentions a fifth variant as of now, YOLOv9-T (tiny). However, currently, only C and E variants are publicly available for both YOLOv9 and GELAN. We chose the lighter architecture that is available, YOLOv9-C accounting for a low computational resource environment. The YOLOv9-C has 962 layers, 51,031,930 parameters (51 M), and 239.0 GFLOPs during training. However, after completion of training, it can be re-parameterized because inference does not require the auxiliary branch, as discussed earlier. Then the model should become 25.3 M parameters, and 102.1 GFLOPs, same as the corresponding GELAN variant.

\subsection{Problem Statement}
Given an image $I$, there are $m+n$ objects of which n are vehicles in the image. The objects are: \{$x_1, x_2, ...,x_m,v_1,v_2,...,v_n$\} where for any i, $x_i$ represents any object and $v_i$ represents any vehicle. We are to detect each vehicle, $v_i$ for all $i = 1,...,n$ into a class or vehicle type from the set of vehicle classes: \{$c_1,c_2,...c_k,...,c_t$\}, where $k$ is the number of vehicle classes considered in this experiment, and $t$ is the total possible number of vehicle classes in the context or region. Each vehicle is detected using a bounding box, described by a tuple: $(x_b, y_b, w_b, h_b)$ where top-left corner, width, and height of the bounding box are $(x_b,y_b)$, $w_b$, and $h_b$, respectively. The model is trained with images with bounding boxes and a class assigned to it from the vehicle classes set. After training, the model can predict bounding boxes around the vehicles for an image $I$. The model meets the requirement if it can detect all $n$ number of vehicles in the image $I$. The performance of the model is not good if it fails to detect all vehicles, meaning it detects less than $n$ number of vehicles, or classifies a wrong class for any vehicle. The number of vehicle classes, $k$ is a design choice of the experiment and depends on the dataset and the vehicle detection objective. 

\subsection{Evaluation Metrics}
We used the metric mean average precision (mAP) to evaluate our models.

\textbf{1) True Positive (TP):} Correct detection of a ground-truth bounding box

\textbf{2) False Positive (FP):} Incorrect detection of a non-existent object or misplaced bounding box of an existing object

\textbf{3) False Negative (FN):} Undetected ground-truth bounding box

In the context of object or vehicle detection, there is no True Negative (TN) since an infinite number of bounding boxes may meet such criteria. Correct and incorrect detection is decided based on Intersection over Union (IoU). 

\textbf{4) Intersection Over Union (IoU):} IoU means Intersection Over Union, is a metric that shows how much area of the ground truth bounding box is overlapping with the predicted bounding box. By considering a threshold value for IoU, correct and incorrect detection is decided. For instance, an IoU of threshold 0.5 is considered in this experiment. So, when the predicted bounding box overlaps at least 50\% with the ground truth bounding box, it is deemed as correct detection, and vice versa. 

\textbf{5) Precision (P) and Recall (R):} Precision means the ability to detect relevant objects i.e., the percentage of correct positive detections. On the other hand, recall means the ability to detect all cases i.e., the percentage of correct positive detections among all ground truths. The formula to calculate precision and recall are,
\begin{equation}
    \text{P} = \frac{ \text{TP} }{ \text{TP} + \text{FP} } = \frac{\text{TP}}{\text{All detections}}
\end{equation}

\begin{equation}
    \text{R} = \frac{ \text{TP} }{ \text{TP} + \text{FN} } = \frac{\text{TP}}{\text{All ground truths}}
\end{equation}

\textbf{6) Mean Average Precision (mAP):} The formula for calculating mAP is,

\begin{equation}
    \text{mAP} = \frac{1}{N} \sum_{i=1}^{N} AP_i
\end{equation}

where, $N$ is the number of classes, $AP_i$ is average precision of class $i$. We chose our model at the checkpoint where the model had the most mAP at IoU of 0.5, denoted as mAP50 or mAP0.5 throughout this report. Average Precision (AP) is the numerical value, the average of precision values across recall values between 0 and 1.  Further understanding of evaluation metrics in the object or vehicle detection task can be referred from \cite{padillaCITE2020}.

\subsection{EXPERIMENTAL SETUP}
We leveraged the transfer learning approach \cite{torrey2010transfer} by using the pre-trained model of the original YOLOv9 on the MS COCO dataset \cite{lin2014microsoft} for weight initialization. As we trained our model for 40 epochs with SGD optimizer \cite{robbins1951stochastic} during different experiments, we stopped mosaic augmentations before the last 5 epochs, since the default was 15 when training 500 epochs on MS COCO dataset in the original paper \cite{wang2024yolov9}. A batch size of 32 and an image size of 640 were used during the experiments. We tuned some of the hyperparameters by observing model performance on the validation set manually using the trial and error method. Experiments showed that hyperparameters: final learning rate (lrf) = 0.01, momentum = 0.85, warmup\_momentum = 0.7, and mixup = 0.2, and keeping the rest as default, provided better results during validation considering mAP-50. 

\section{Results}
\label{sec:results}
\begin{figure}
\centering
\includegraphics[width=3.5in]{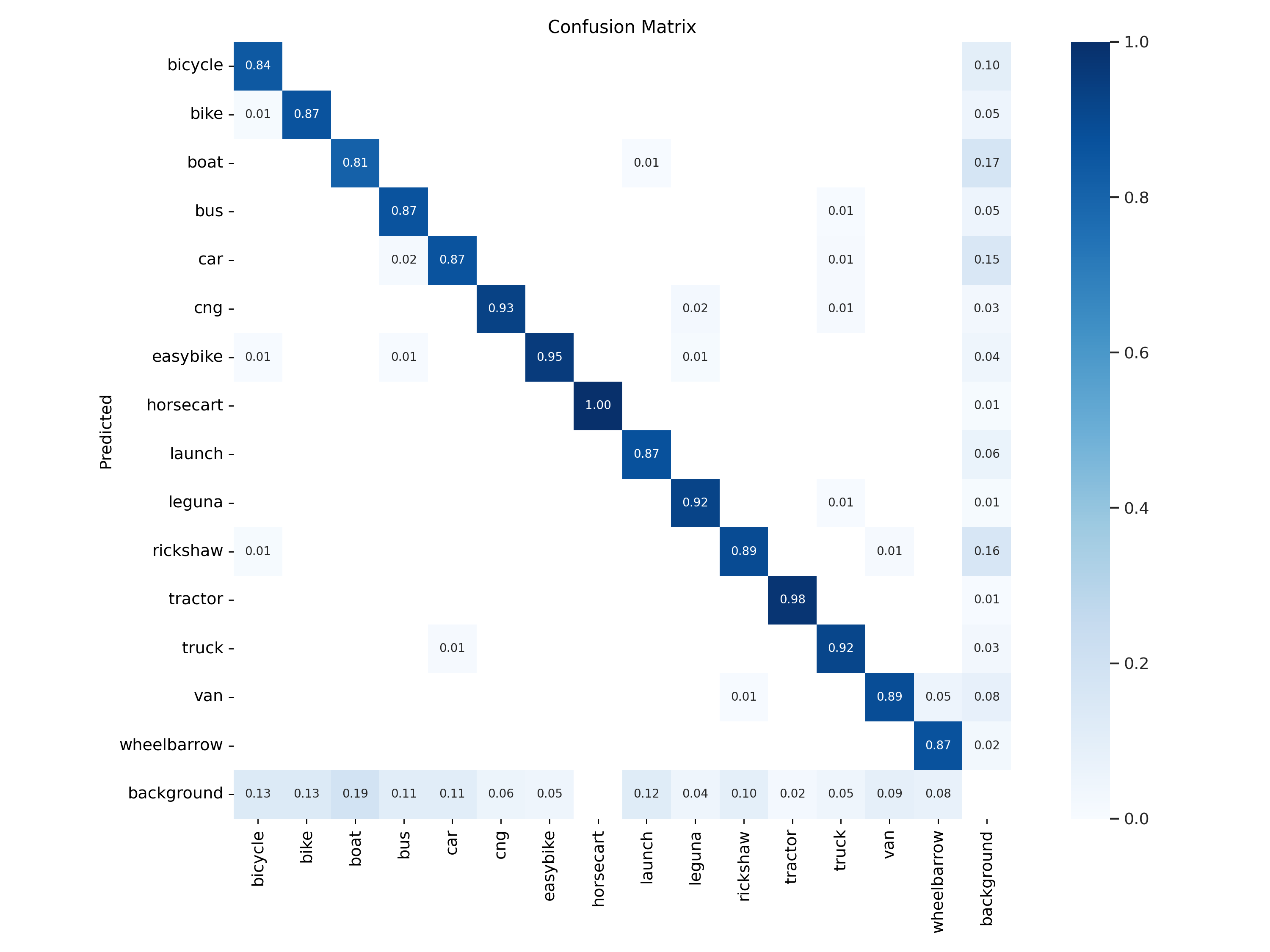}
\caption{Confusion matrix of our fine-tuned YOLOv9 vehicle detection model for the test set.}
\label{fig_confusion}
\end{figure}

\begin{figure}
\centering
\includegraphics[width=3.5in]{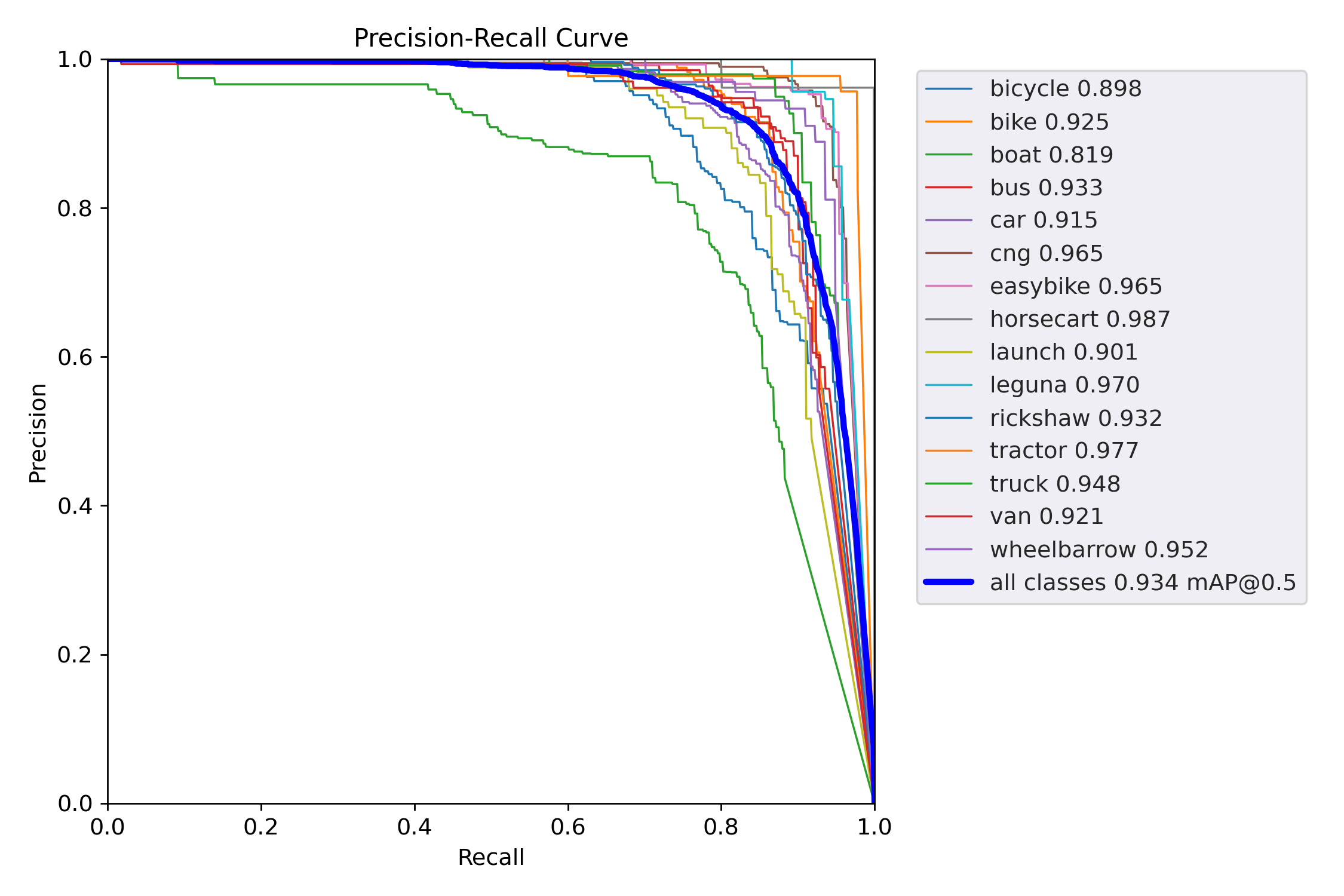}
\caption{Precision-recall diagram of our model on test set.}
\label{fig_PR}
\end{figure}

Table~\ref{tab_res} shows the results of our fine-tuned model on the test set. The table shows instances or number of times a vehicle is seen in the images in the testing dataset, precision, recall, and mAP values for each class of vehicles. There were a total of 908 images in the test set. We see that our model achieved an mAP of 0.934 on average for all classes when the IoU of 0.5 threshold is considered. Even from the 0.5-0.95 IoU threshold, our model obtained an mAP of 0.732 on average, meaning the predicted bounding boxes have a fair amount of overlapping region with ground truth. It is observed that the lowest mAP50 was obtained for the class boat. Through hand checking on the predicted outputs of our model on boat class images from testing data, we saw that some boats were not detected for an image that had boats a bit far, also in bird's eye view. However, the main reason for the low mAP for the boat is due to confusion of this class with the background. We can understand such false detections further if we look at the confusion matrix on the test set in Figure \ref{fig_confusion}. The model showed the highest confusion with the background for the class boat. Our interpretation is that, usually the background of boat images is water, such a background is not common for other classes. Thus the model is less robust for the detection of boats or vehicles on water. We also observe that some wheelbarrows were predicted as vans, as they look similar. Other notable false detected pairs are predicting buses as cars, and legunas as CNGs. Finally, Figure \ref{fig_PR} shows the precision-recall graph of our model on the test set. A color is used more than once because of many classes. The area under the curve on this diagram is the mAP value. The green curve representing the boat class, is the smallest, as it has the least mAP. The highest mAP is observed for the horsecart class with the ash curve, observed as the biggest in this diagram. The weighted blue curve in the illustration is the all-classes average precision-recall curve at 0.5 IoU, area of which is the main evaluation metric of our experiment, mAP-50 = 0.934. 

\begin{table}
\begin{center}
\caption{Testing set results of proposed YOLOv9 model.}
\label{tab_res}
\begin{tabular}{lrllll}
\hline
        \textbf{Class} & \textbf{Instances} & \textbf{P} & \textbf{R} & \textbf{mAP50} & \textbf{mAP50-95} \\
\hline
\text{all} & 2730 & 0.921 & 0.862 & 0.934 & 0.732 \\
\text{bicycle} & 194 & 0.896 & 0.755 & 0.898 & 0.61 \\
\text{bike} & 225 & 0.939 & 0.827 & 0.925 & 0.717 \\
\text{boat} & 273 & 0.835 & 0.714 & 0.819 & 0.502 \\
\text{bus} & 171 & 0.938 & 0.842 & 0.933 & 0.724 \\
\text{car} & 323 & 0.898 & 0.82 & 0.915 & 0.694 \\
\text{cng} & 222 & 0.95 & 0.919 & 0.965 & 0.802 \\
\text{easybike} & 173 & 0.931 & 0.931 & 0.965 & 0.794 \\
\text{horsecart} & 25 & 0.942 & 1 & 0.987 & 0.79 \\
\text{launch} & 134 & 0.87 & 0.821 & 0.901 & 0.683 \\
\text{leguna} & 93 & 0.961 & 0.892 & 0.97 & 0.848 \\
\text{rickshaw} & 370 & 0.902 & 0.846 & 0.932 & 0.694 \\
\text{tractor} & 45 & 0.956 & 0.956 & 0.977 & 0.879 \\
\text{truck} & 170 & 0.937 & 0.888 & 0.948 & 0.766 \\
\text{van} & 235 & 0.923 & 0.864 & 0.921 & 0.742 \\
\text{wheelbarrow} & 77 & 0.943 & 0.861 & 0.952 & 0.727 \\
\hline
 \end{tabular}
\end{center}
\end{table}

\begin{figure*}
\centering
\includegraphics[width=4.8in]{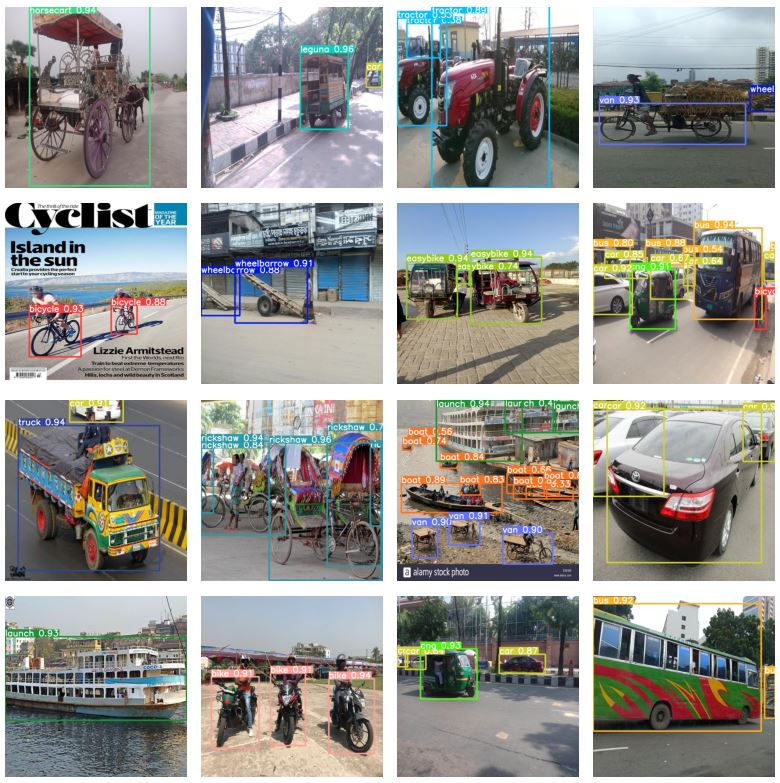}
\caption{Inference samples on test set.}
\label{figyolov9detect}
\end{figure*}

Figure \ref{figyolov9detect} shows test data samples from each classes and one from the mixed randomly chosen. 16 images from test set results are visualized randomly in this Figure, with one image from each class type and one from the mixed. It performs well in a congested environment (column 4 from the left of the second row), showing applicability during traffic congestion on the road. 

We also tested our model on our own recorded video on the streets of Dhaka, and observed competitive detection precision from our model. During inference on video, the average speed was 0.2 ms for pre-processing, 12.6 ms for inference, and 1.1 ms NMS (non-maximum suppression) for every image of shape 640. The model missed some detection for micro-sized vehicles at a very large distance. It may make an error while detecting a vehicle class with a changed form that it has not seen before, such as a loaded van with water cans (goods). Instances of detection results on our recorded video are included in the Appendix.

\section{Discussions}
\label{sec:discussions} 
\subsection{Interpretation of Results}

Although a vehicle detection system can be implemented through other means, such as sensors \cite{kessler2023detection}, hardware, and radars apart from image/video from cameras, a vision based method is better in terms of flexibility and cost \cite{yang2018vehicle,gholamhosseinian2021vehicle}. Researchers in \cite{toha2021dhakanet} argued that vehicle detection systems for the roads should have relatively lightweight architectures because these systems typically need to be deployed on platforms having constrained computing resources, such as embedded systems and edge devices \cite{kulkarni2020real} on cameras. We propose our model to be deployed on CCTVs on the streets to detect vehicles, creating a detection vehicle system in the city. A lighter weight architecture is a research question in the task of object detection \cite{wang2023ma}. So, we performed experiments using the available lighter-weight architecture, YOLOv9-c. However, when lighter variants in the lineup or the lightest one, YOLOv9-s are released, further experiments should be done to perform sensitivity analysis to find the optimal variant considering speed and accuracy when deploying. The advantage of the YOLO series has always been the high FPS (frame per second) over other models keeping reasonable high accuracy. We did not test our model on video to evaluate this scope, but this also leaves for future experiments before practical implementation according to the deployment environment.

The performance of model may fluctuate with weather changes, e.g. darkness, fog, rain, partial occlusion, and blurring. To improve the model while developing an actual application, larger datasets with more variations in conditions should be incorporated. For instance, \cite{zhou2023all} showed how to use their novel framework Illumination-Adjustable GAN to generate synthetic night-time data from day-time data for robustness. In our experiment, ill-weathered training samples may not be enough in our study to detect vehicles during extreme weather events, like, urban flooding due to intense rain, frequently occurring in Dhaka \cite{rahman2020future}. We argue it may be difficult for the model to learn high-level features of the target vehicle class when some portion of the vehicle is underwater, such as wheels. Another suggestion for improving during actual deployment is to add null class images to the dataset. Null class image is an image with no vehicles in it, so it is an image $I$ with objects: \{$x_1, x_2, ...,x_m$\}. If the model is deployed on CCTV, it may get video instances where there is no vehicle on the road. The current model may not infer well because it was never trained on null class images. Again, the model may face with newer vehicle class, $v_{t+1}$ during implementation it was not trained on. It can be made robust in such scenario through zero-shot learning \cite{bansal2018zero}. The dataset is originally based in Dhaka, Bangladesh, but we have observed that it contains some images that are not in the context of Dhaka with different backgrounds, the effect of which during actual operation of the system may be negligible but leaves room for improvement for the dataset. Also, the used original dataset of this paper \cite{tabassum2020poribohon} itself has 20\% images augmented. When this dataset is split into train-valid-test or train-valid sets, the valid and test sets always may get some images that are augmented with training samples. This may result in an overestimated performance for any experiments done using this dataset during validation and/or testing. Testing the model on own collected video data (see Appendix), we assert that the model needs to be further tested for more diverse situations with challenging backgrounds, such as intensely traffic congested roads (e.g., with vehicles, food-carts), varying form of vehicles (e.g., loaded vans with goods), etc. Therefore, developing a more standardized dataset is a research need for the future.


\subsection{Comparison with Recent Studies}\label{subsec:comparison} 

Table~\ref{tab_comparison} compares our proposed model with models from recent works that incorporated a dataset based in Dhaka, Bangladesh. All works in the table are based on finetuning the corresponding YOLO model. We considered contributions that showed performance regarding vehicle detection. A direct comparison may not be fair because each work has been performed in different experimental environment with different datasets consisting of different number of classes, but this table gives an overview of recent progress in this regard, nonetheless. In the dataset column, P-BD is the Poribohon-BD dataset \cite{tabassum2020poribohon} and D-AI is the Dhaka-AI dataset \cite{DVN/POREXF_2020}. Some values in \#Params. column are marked with an asterisk (*), which means they were not explicitly mentioned in the reference, but written from corresponding original YOLO papers matching the architecture variant. mAP is considered at IoU of threshold 0.5. There was no work found using detection algorithms other than YOLO to finetune on a Bangladesh-based dataset with notable accuracy.

\begin{table}
\scriptsize
\begin{center}
\caption{Comparison of our proposed fine-tuned YOLOv9-C model with previous works on Dhaka (or Bangladesh) based dataset. }
\label{tab_comparison}
\begin{tabular}{c|c|r|l|r|l}
    \textbf{Ref.} & \textbf{Dataset} & \textbf{\#Classes} & \textbf{Model} & \textbf{\#Param.} & \textbf{mAP} \\    
\hline
   [Ours] & P-BD & 15 & YOLOv9C & 25.3 M & \textbf{0.934}\\ 
   \cite{salekin2023bangladeshi} & P-BD & 15 & YOLOv8S & \textbf{11.2 M} & 0.913 \\  
   \cite{munir2023vehicle} & D-AI & 21 & YOLOv7 & *36.9 M & 0.512 \\
   \cite{hossain2023real} & Custom PMB & 8 & YOLOv7X & *71.3 M & \textbf{0.968} \\   
   \cite{sarker2023bangladeshi} & Custom & 20 & YOLOv7 & 37.2 M & 0.802 \\ 
   \cite{rahman2022densely} & D-AI & 21 & YOLOv5X & *86.7 M & 0.458 \\  
   \cite{rafi2022performance} & Custom & 21 & YOLOv5L & *46.5 M & 0.669 \\  
   \cite{alamgir2022performance} & D-AI+P-BD & 21 & YOLOv5L & 87.7 M & 0.287 \\  
   \cite{daspassphrase2021} & D-AI & 21 & YOLOv5L+X & - & 0.895 \\
   \cite{bari2021performance} & D-AI & 21 & YOLOv3 & *65.3 M & 0.76 \\      

\end{tabular}
\end{center}
\end{table}

It is observed that our model relatively outperforms previous finetuned models. Although \cite{hossain2023real} has higher mAP than this paper, it is worth mentionable that their model is solely based on a single location, Padma Multipurpose Bridge (in short, PMB) in Dhaka, and only trained and evaluated on 8 classes of vehicles from their own developed custom dataset naming it as PMB Vehicle Image Dataset, relevant to that location only. However, including the custom PMB dataset, none of the custom datasets in Table \ref{tab_comparison} were found to be publicly available for future research. Next, we have not compared the models for speed since each of these models has been trained in different environments, exporting methods, and versions. Thus, a direct comparison may fail to provide a fair latency comparison. Currently, the number of parameters, \#Params. (M) is lowest for \cite{salekin2023bangladeshi}. In the future, with the release of the YOLOv9-S model that has 7.1 M parameters, may be trained to obtain better results in terms of parameters, since the YOLOv9-S already showed higher mAP with lower parameters than the corresponding YOLOv8-S with 11 M parameters for the MS COCO dataset, according to the original YOLOv9 work \cite{wang2024yolov9}.

\subsection{Proposed Technique to Process Model Output}

\begin{figure}
\centering
\includegraphics[width=2.6in]{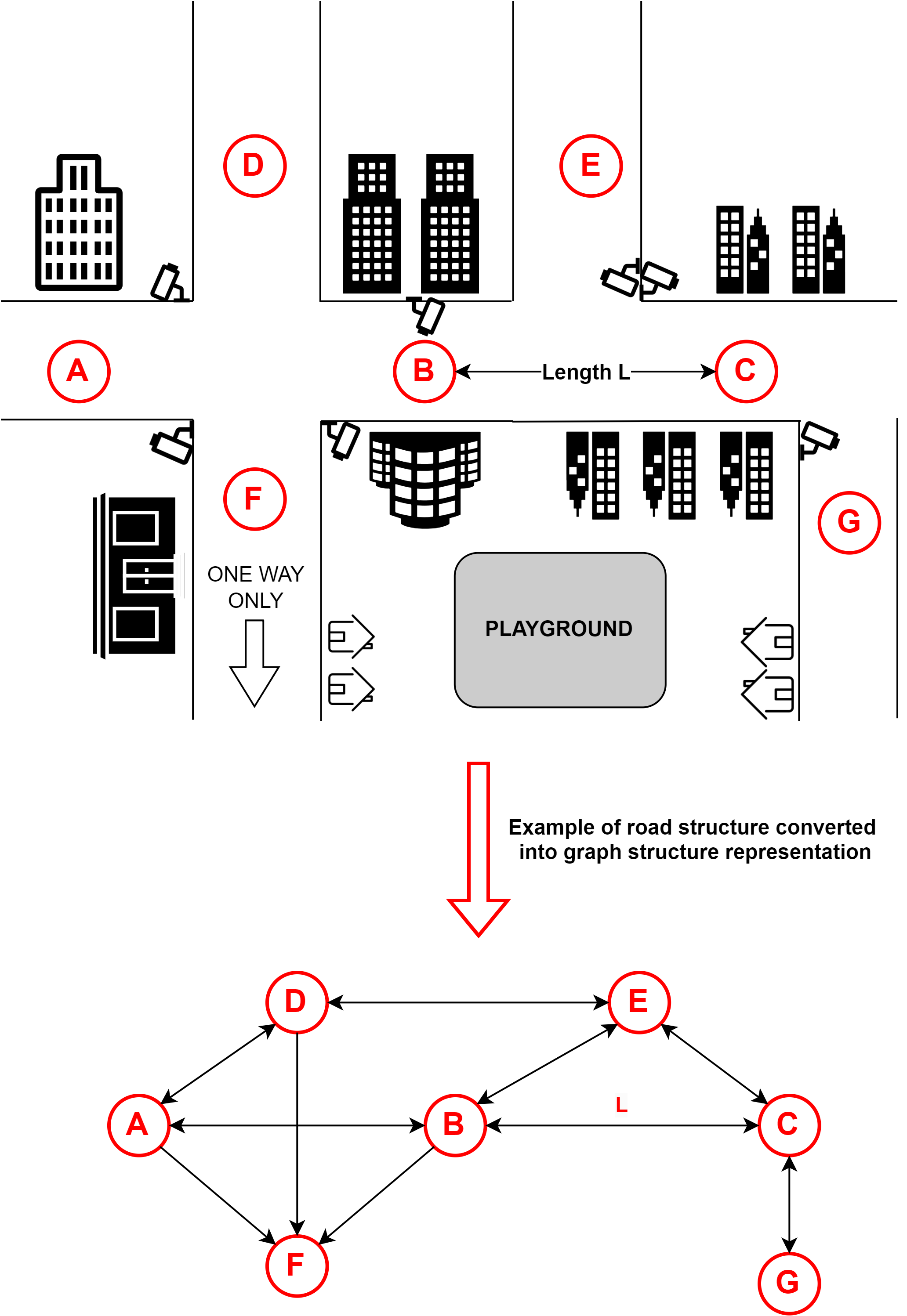}
\caption{Road structure converted into graph representation to process traffic data obtained from vehicle detection system deployed in CCTVs in the city.}
\label{fig_map}
\end{figure}

We propose a technique on how to process the vehicle detection output data for further uses. As shown in Figure \ref{fig_map}, after setting up CCTV on roads and deploying our vehicle detection model, each CCTV will provide information on the type and frequency of vehicles that pass through the road, like \cite{tian2015accurate}. Let us consider CCTVs are placed at various locations of different streets from different angles converting all lanes of the road. Each CCTV can output number of vehicles with identified type in real-time as temporal data. We can leverage the concept of graph structure to represent the traffic flow information of the roads from the cameras deployed with our model. We represent the roads of the city in a bi-directed graph, where each node will represent the point where a CCTV is located and detects vehicles. The link may represent the distance between the two nodes or CCTVs, i.e., L between nodes B and C. Direction shows vehicle flow direction. See that a one-way road has only one direction when transformed into a graph network. Thus, from different CCTVs on different positions of the roads, we obtain real-time graph network data where each node gives traffic flow information of the respective position. We may visualize and find spatiotemporal trends by analyzing model outputs from cameras through this representation. Thus, a large-scale traffic information data collection system can be developed, leading to traffic flow prediction \cite{li2022spatio,zhong2023estimating}, surveillance \cite{wang2020robust}, and monitoring in the whole city \cite{yang2018vehicle}. This will help to understand how people move within the city during extreme weather events or emergency situations, such as COVID-19, extreme rainfall, or disasters. Transportation authorities can make better-informed decisions to handle crises.

\subsection{Applications of Our Model in ITS for Benefits in Society}

\begin{figure*}
\centering
\includegraphics[width=\textwidth]{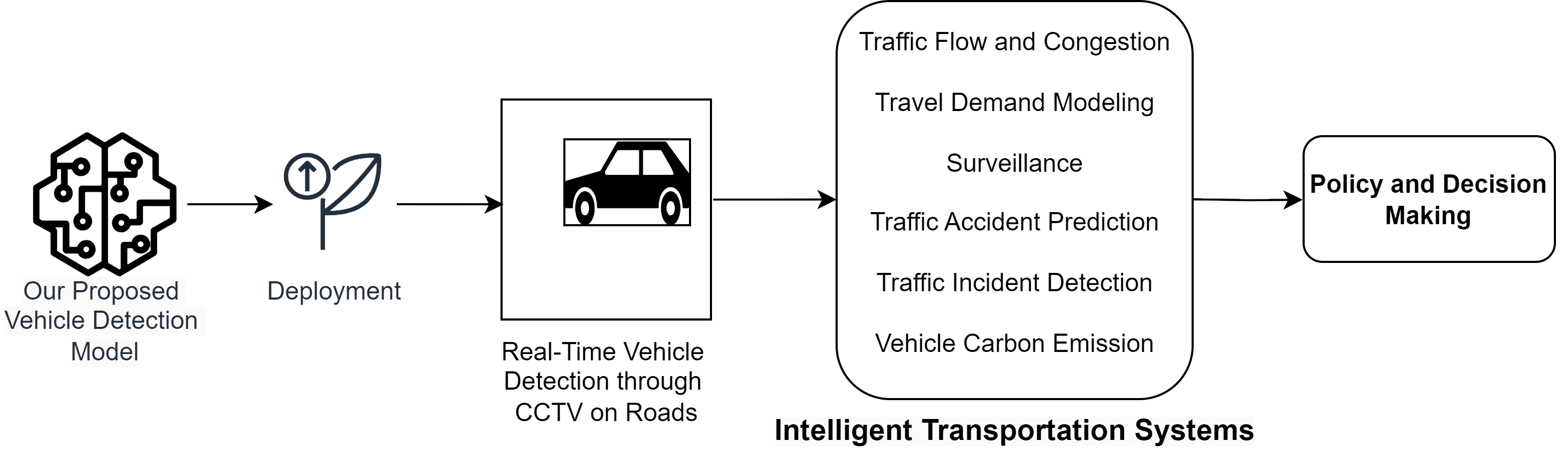}
\caption{Proposed framework towards the Intelligent Transportation Systems applications based on vehicle detection model.}
\label{fig_applications}
\end{figure*}

Here, we shed light on potential ITS applications that can be developed on top of our vehicle detection model after using our proposed technique, to show how it can solve complex transportation problems realizing ITS. This will provide a rationale for implementing our proposed vehicle detection system throughout the city to gain societal benefits. Figure \ref{fig_applications} shows a framework for potential ITS applications that we can develop based on our model. Our model can be used as a foundation for further algorithm developments for solving ITS research questions but not limited to, modeling traffic flow and congestion, travel demand modeling, surveillance, predicting traffic accident risks on the map, detecting incidents on roads, and estimating vehicle carbon emission to tackle global warming. After analyzing the results, the authority can make informed policy and decision-making. 

We will elaborate on some key research questions that can leverage the vehicle detection system in the city. One study employed vehicle detection to create a system for advance warning of congestion \cite{chintalacheruvu2012video}. By incorporating street accident data through annotating accident locations into the nodes and links of our graph representation for example, traffic accident risks can be predicted, as done in the study \cite{hu2021research}. Another contribution \cite{chen2022prediction} showed how YOLO based vehicle detection can be utilized to create "vehicle density maps" from mixed traffic flow to predict optimal timing of traffic signals at road intersections. Another application may be incident detection e.g., detecting a vehicle staying in a non-parking area for a long time, like \cite{elassy2024intelligent}. By identifying vehicle types with temporal variation and geographical location, future mobility demand studies can be done. Observations on patterns of movement of people and kinds of vehicles moving through different roads can infer congestion patterns and help to enact necessary policy accordingly, solving the city's mobility issues. Global warming is one of the major challenges in the 21st century, and estimating carbon emission is a heavily prioritized research question \cite{wang2017method}. The mobility data obtained from the vehicle detection system across the city can assist in estimating carbon emissions from vehicles. Research may be done to assess environmental impact assessment (e.g., noise and air) considering the output data of the detection model. Overall, policy-making to optimize the transportation system based on the mobility data created through the vehicle detection system will increase the quality of life of city residents. The informed decision-making for effective traffic management will ensure economic benefits (e.g, energy saving, reduced accidents, and reduced costs), environmental benefits (e.g., reduction of air and noise pollution, reduced harmful gas emissions), and social welfare (e.g, better travel and mobility, safety, time-saving, and high quality of life due to better transportation) \cite{dilek2023computer,ravichandran2023efficient}. Finally, although such a system can solve many challenges in society, it may also create ethical and privacy issues \cite{waelen2023ethics}. Hence, guidelines should be developed side by side to handle ethical concerns.

\section{Conclusions}\label{sec_conclusions}
The growing need for mobility and its effects, including traffic jams, inefficiencies, and increased risk of accidents in the transportation sector, has given rise to the need for intelligent transportation systems. Detecting vehicles using AI and deep learning can help solve many transportation issues in Dhaka, Bangladesh by realizing ITS. This paper is the first to address the problem of vehicle detection task by fine-tuning the YOLOv9, one of the most prominent object detection techniques. Results show that the model achieved an mAP of 0.934 at the IoU threshold of 0.5 for a dataset based in Dhaka, Bangladesh. Comparison with previous finetuned models shows it has state-of-the-art performance in detecting Bangladeshi vehicles. Future works include training the model on images with null class, more diverse challenging backgrounds and environments to make the model more robust, and developing a standardized dataset for this task. The model should be deployed on cameras to create a real-time vehicle detection system in the city. The proposed technique in this paper to handle model output traffic data will facilitate building algorithms on top of the model to work on more advanced research questions in ITS. Discussions on potential applications of our proposed vehicle detection model will guide relevant authorities to understand the scope of societal benefits. This contribution will play a vital role in shaping the transportation infrastructure of Dhaka, Bangladesh into an intelligent one to realize the "Smart Bangladesh Vision 2041".

\section*{Acknowledgment}
Thanks to 1) Ragib Amin Nihal for insightful discussions and feedback on the final draft, 2) M.M. Harussani for feedback on the final draft, and 3) Nafisa Tasnim for capturing and providing the data used in graphical abstract Figure \ref{fig_abstract} and Appendix respectively.

\section*{Appendix}
\subsection*{Detection results on our video}

\begin{figure}[H]
    \centering
    \includegraphics[width=2.4in]{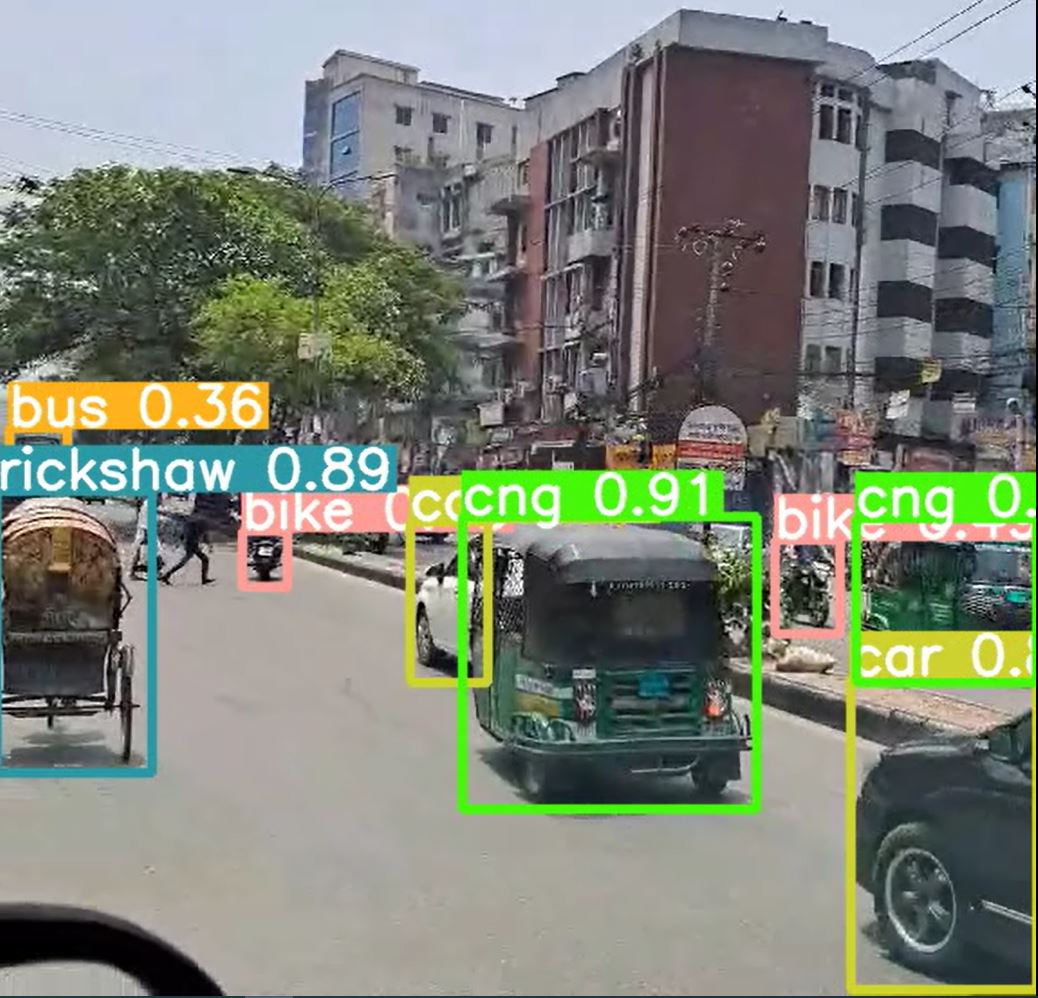}
    \caption{An instance of our video showing good detection for vehicles distanced far.}
    \label{fig:ap1}
\end{figure}

\begin{figure}[H]
    \centering
    \includegraphics[width=2.4in]{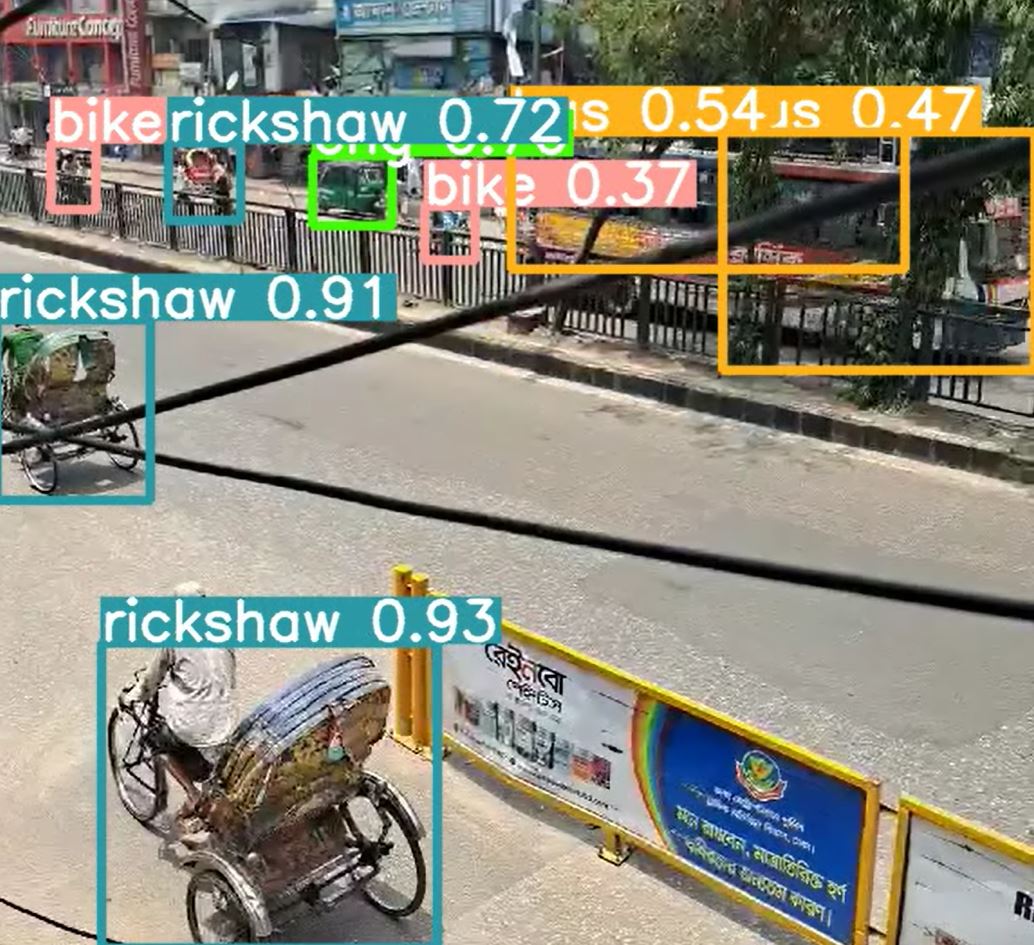}
    \caption{An instance of our video where predicted the bike with 0.37 confidence score is actually a van that is loaded with water cans. One micro-sized vehicle beside the detected bike on top-left was not detected.}
    \label{fig:ap2}
\end{figure}

\bibliographystyle{unsrt}  
\bibliography{references}

\end{document}